\documentclass[letterpaper, 10 pt, conference]{ieeeconf} 

\usepackage{amsmath,amssymb,amsfonts}
\usepackage[ruled,linesnumbered]{algorithm2e}
\usepackage{graphicx}
\usepackage{textcomp}
\usepackage{xcolor}
\usepackage{multicol}
\usepackage{multirow}
\newsavebox{\tablebox}
\usepackage{ulem}
\usepackage{makecell}

\IEEEoverridecommandlockouts                     % This command is only needed if 
\SetAlFnt{\footnotesize}

\PassOptionsToPackage{hyphens}{url}\usepackage{hyperref}

\newcommand{\sig}[1]{{\small\textsf{{#1}}}}

\newcommand{\Comment}[1]{}

\begin{document}

\title{ComOpT: Combination and Optimization for Testing \\ Autonomous Driving Systems}

\author{Changwen Li$^{1,5\#}$, Chih-Hong Cheng$^{2\#}$, Tiantian Sun$^{3}$, Yuhang Chen$^{4,5}$, and Rongjie Yan$^{1,5}$
\thanks{$^{1}$ State Key Laboratory of Computer Science, ISCAS, China}
\thanks{$^{2}$ Independent contributor, Germany } 
\thanks{$^{3}$ Beijing University of Technology, China} 
\thanks{$^{4}$ Technology Center of Software Engineering, ISCAS, China}
\thanks{$^{5}$ University of Chinese Academy of Sciences,  China}
\thanks{$^{\#}$ The first two authors are listed in reverse alphabetical order but they contribute equally. Chih-Hong Cheng works on this project as his hobby  activities and thereby voluntary and not-for-profit; opinions stated in this paper shall not be viewed as an official statement from his organization. }
\thanks{$^{*}$ Correspondence to yrj@ios.ac.cn, cheng.chihhong@gmail.com}
\vspace{2mm}
\\
\url{https://github.com/safeautonomy/ComOpT}
}

\maketitle

\begin{abstract}
 ComOpT is an open-source research tool for coverage-driven testing of autonomous driving systems, focusing on planning and control. Starting with (i) a meta-model characterizing discrete conditions to be considered and (ii) constraints specifying the impossibility of certain combinations, ComOpT first generates constraint-feasible abstract scenarios while maximally increasing the coverage of $k$-way combinatorial testing. Each abstract scenario can be viewed as a conceptual equivalence class, which is then instantiated into multiple concrete scenarios by (1) randomly picking one local map that fulfills the specified geographical condition, and (2) assigning all actors accordingly with parameters within the range. Finally, ComOpT evaluates each concrete scenario against a set of KPIs and performs local scenario variation via spawning a new agent that might lead to a collision at designated points. We use ComOpT to test the Apollo~6 autonomous driving software stack. ComOpT can generate highly diversified scenarios with limited test budgets while uncovering problematic situations such as inabilities to make simple right turns, uncomfortable accelerations, and dangerous driving patterns. ComOpT participated in the 2021 IEEE AI Autonomous Vehicle Testing Challenge and won first place among more than~$110$ contending teams. 
\end{abstract}

\section{Introduction}

The development of autonomous driving (AD) technologies has reached the stage where the safety of such systems is a dominating factor for defining success. For verification and validation of autonomous vehicles in a fixed operational design domain (ODD), simulation-based testing is one of the highly recommended methods for modular testing of planning and control systems. The simulation environment can provide object labels such as bounding boxes (as a replacement of the perception module), allowing the prediction and planning modules to be tested in isolation. Nevertheless,  the critical challenge remains to be designing the \emph{test case generation and management} module. Within a limited test budget, the test case generation module should \emph{outsmart the AD system under test} by creating scenarios that lead to undesired behavior (e.g., collision). Simultaneously, to demonstrate sufficient coverage over the ODD, the generation of test cases should be \emph{coverage-driven} while ensuring \emph{diversity}. 

Towards the aforementioned challenges, we present ComOpT, an open-source research tool (under the AGPL v3 license) for coverage-driven testing of autonomous driving systems. As of September 2021, ComOpT interfaces to the open-source simulator LGSVL~\cite{rong2020lgsvl} and the Baidu Apollo autonomous driving software stack~\cite{BaiduApollo}. Internally, ComOpT integrates an \emph{axiomatic approach} to generate \emph{abstract scenarios} from a pre-defined list of categories. Every element in a category has its physical interpretation reflected in the simulation environment. One combination forms an abstract scenario, which can be instantiated in the simulation environment by (a) assigning a local map as well as (b) concretizing every element by picking a value in its associated parameter range. Although the axiomatic approach of abstract scenario generation sounds intuitive at first glance, it suffers from three crucial limitations in terms of realization: 

\begin{enumerate}
    \item \emph{Combinatorial explosion}: Given $N$  categories with each category having only~$2$ elements, there exist~$2^N$ possible abstract scenarios  in the worst case.
    \item \emph{Feasibility considerations}: Certain combinations are semantically unclear (e.g., making a \sig{left-turn} in a \sig{straight-line} road segment) or simply  nonexistent in the ODD under consideration (e.g., no roundabout with pedestrian crossings in the map).
    \item \emph{Semantically enriched map}:  Concretizing an abstract scenario requires searching for high-level concepts such as intersection types; these concepts are nonexistent in standard map formats.

\end{enumerate}

For the first two issues, ComOpT utilizes the technique of \emph{constraint-based $k$-way combinatorial testing}~\cite{cheng2018quantitative,cheng2019nn}. Combinatorial testing~\cite{nie2011survey}  provides a coverage metric by requiring test cases to cover all element tuples for any~$k$ categories.
The constraint-based variation allows integrating feasibility constraints, while implementing it using optimization solvers (e.g., mixed-integer linear programming) allows suggesting new test cases that maximally increase coverage. For the third issue, ComOpT implements a special module that allows querying high-level semantic information. In particular, it allows searching for local maps that can satisfy a specific combination of conditions (e.g., find all T-way junctions having pedestrian-crossings).

Finally, given a concrete scenario tested against a set of KPIs, ComOpT includes methods that can perform \emph{local perturbation} over the scenario based on an innovation called \emph{agent spawning}. Concretely, ComOpT introduces new agents (e.g., vehicles) to challenge the AD system. How these agents are introduced depends on the high-level behavioral pattern of the ego vehicle extracted from the simulated trace.

ComOpT participated in the 2021 IEEE AI Autonomous Vehicle Testing challenge. Apart from  convincing scenario coverage and diversity metrics, ComOpT discovered numerous undesired scenarios for the Apollo Autonomous Driving SW stack, including situations to hit crossing pedestrians or inabilities to make trivial right turns. Ultimately, ComOpT won \emph{first} place in the competition (within more than~$110$ contending teams), serving as objective evidence of the system performance.

\section{Related Work}

Testing autonomous driving systems has been an active research field, and  many autonomous driving safety standards such as ISO PAS 21448 SOTIF (Safety of Intended Functionality)~\cite{SOTIF} or UL~4600~\cite{koopman2019safety} consider simulation-based testing to be instrumental. There are recent programmatic methods such as Scenic~\cite{fremont2019scenic} or Paracosm~\cite{majumdar2021paracosm} that allow specifying and generating scenarios for AD testing. The abstraction that ComOpT takes is one layer higher, supported by an automatic translation from the map of the ODD to a list of sub-maps matching the specification. This implies that given a specific ODD (e.g., Munich city), methods suggested by Scenic or Paracosm cannot easily provide coverage claims over the ODD. These methods are more applicable to re-specify a small number of scenarios, such as known crash scenarios. The Scenic tool, together with the CPS falsification tool VerifAI~\cite{dreossi2019verifai}, also participated as an integrated testing framework in the 2021 IEEE AI testing competition~\cite{viswanadha2021addressing} but  failed to compete against ComOpT.\footnote{For the complete evaluation criteria, we refer readers to the IEEE AI AV testing competition website \url{http://av-test-challenge.org/}}

Within the scope of autonomous driving, prior work on combinatorial testing focuses on testing deep neural networks~\cite{cheng2018quantitative,ma2018combinatorial,cheng2019nn}. However, the black-box feature also makes it also highly applicable to be used in testing prediction and planning modules, as demonstrated by this work. With the semantic enrichment of maps, we can use combinatorial testing to argue the completeness of abstract scenarios \emph{against the ODD}, thereby further differentiating ComOpT from other works. Apart from an \emph{axiomatic approach} demonstrated by ComOpT, other commonly seen approaches include replaying existing crash scenarios (e.g., NHTSA accident database~\cite{thorn2018framework}) or scenario databases created by physical driving or by collective efforts (e.g., Pegasus~\cite{zlocki2017database} or SafetyPool~\cite{safetypool}). One can run these scenarios (from standard formats ASAM OpenScenario~\cite{OpenScenario}) and subsequently, perform scenario variations via various CPS falsification techniques~\cite{annpureddy2011s,ben2016testing,koren2018adaptive,dreossi2019verifai,zhang2020constraining,barbot2020falsification}. While we surely understand the benefit of scenario replay, ComOpT aims more - scenario replay and variation are incapable of arguing diversity and completeness. Finally,  the previously mentioned CPS falsification techniques~\cite{annpureddy2011s,ben2016testing,koren2018adaptive,dreossi2019verifai,zhang2020constraining,barbot2020falsification} largely take low-level information traces such as vehicle position and velocity profiles and perform the search. In contrast, our scenario variation method (agent spawning) also integrates high-level semantic information such as behavioral patterns of the ego-vehicle and positions related to lanes to avoid performing a search in unnecessarily high parameter dimensions. Some of the low-level scenario variation  methods also integrate constraint-free combinatorial testing in dealing with discrete variables~\cite{annpureddy2011s,majumdar2021paracosm}. However, ComOpT uses a constraint-based version on the abstract (high-level) scenario generation; constraints are introduced due to considerations on encoding scenario infeasibility. 

\section{Inside ComOpT}\label{se:method}

\begin{figure}[h]
\includegraphics[width=0.9\columnwidth]{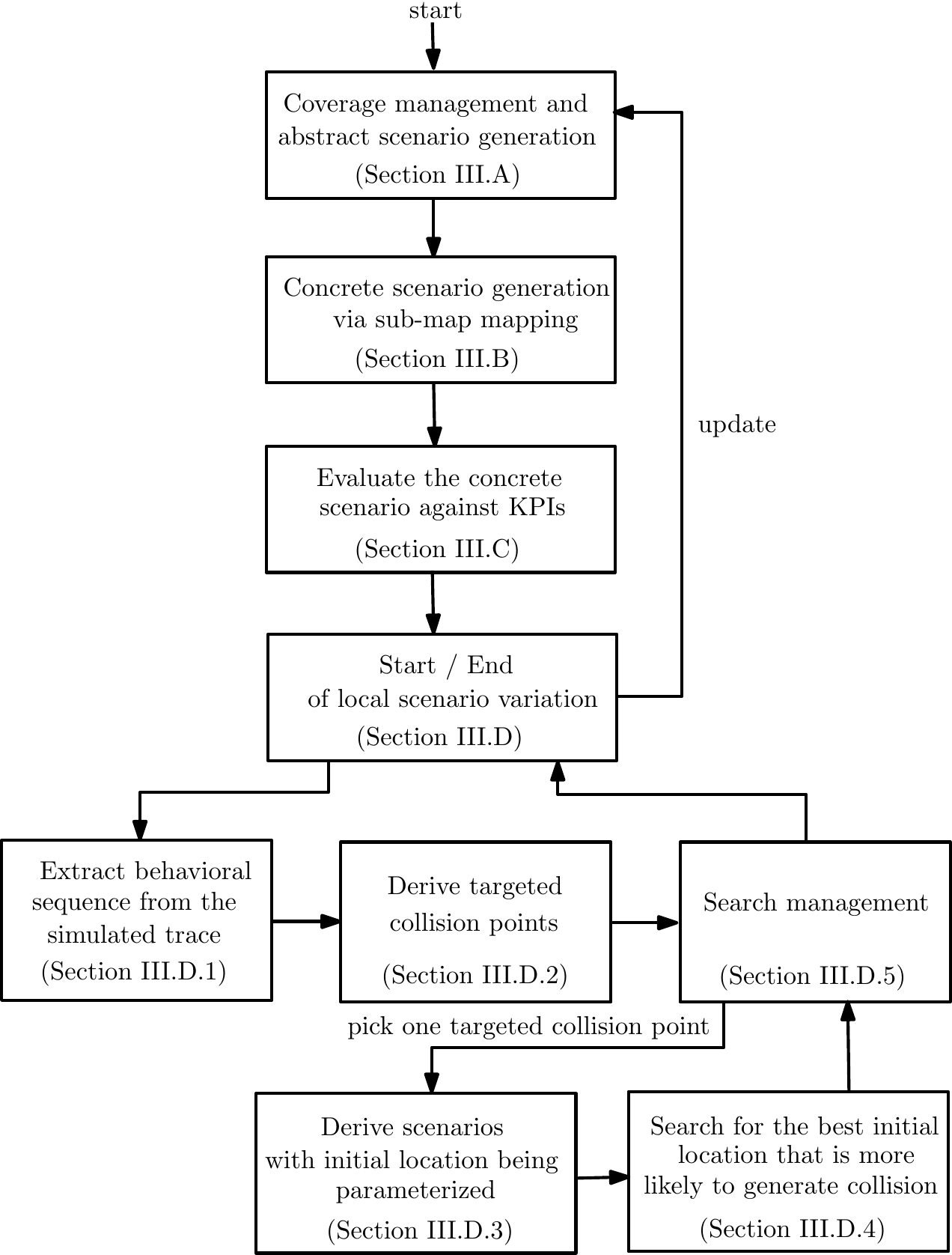}
\centering
\vspace{-2mm}
\caption{The underlying workflow of ComOpT}
\label{fig:workflow}
\end{figure}

In this section, we detail the underlying techniques integrated inside ComOpT. To ease understanding, Figure~\ref{fig:workflow} provides a summary of the workflow and associates each function with the corresponding subsection in this paper. 
   
\vspace{-4mm}

\subsection{Abstract scenario generation via combinatorial testing}  

In this stage, the input of ComOpT is a list of discrete categories serving as the basis for abstract scenario generation. To ease understanding, throughout this paper, we consider the following overly simplified three categories. 

\begin{itemize}
    \item $\sig{weather}\in  \{\sig{sunny, rainy, cloudy}\}$,
    \item  $\sig{road}\in  \{\sig{straight}, \sig{T-shaped}\}$, and
    \item    $\sig{ego-action}\in  \{\sig{drive-straight}, \sig{left-turn}, \sig{u-turn}\}$.
\end{itemize}
   
Apart from categories, ComOpT also takes explicitly stored \emph{constraints} stating that certain combinations are impossible\footnote{It is possible to add additional constraints on the fly where if some abstract scenarios are not realizable, then block the further generation of such abstract scenarios using constraint assignments.}. For example, if the road is straight, it is impossible for the autonomous vehicle to take a left turn. Utilizing the defined categories, the statement can be written as a logical formula $ ``\sig{road.straight} \rightarrow \neg\sig{ego-action.left-turn}"$.

\begin{figure}[t]
\includegraphics[width=0.9\columnwidth]{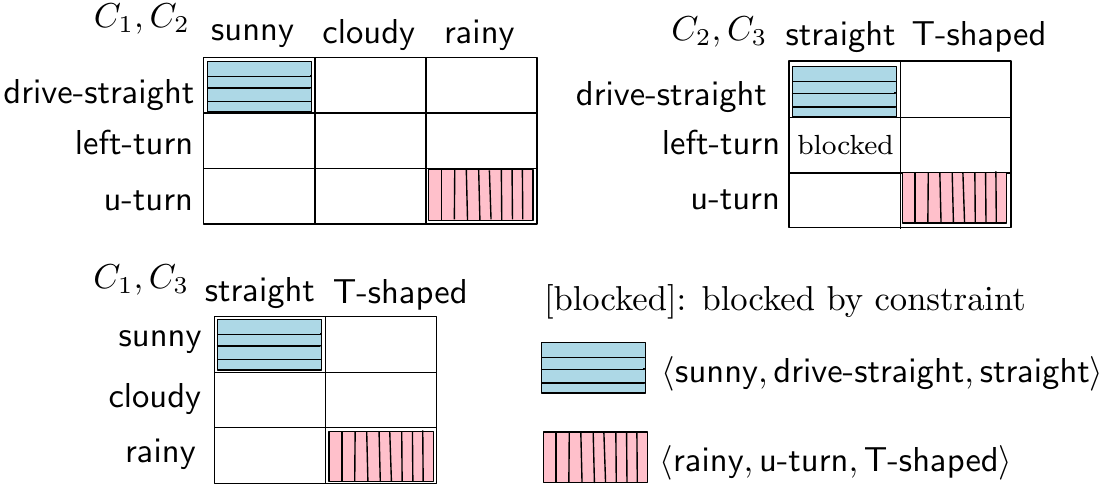}
\centering

\caption{Generating the 2nd scenario $\langle \textsf{rainy}, \textsf{u-turn}, \textsf{T-shaped}\rangle$ maximizing the 2-way projection coverage}
\label{Figure2Projection.coverage}
\end{figure}

ComOpT uses $k$-way combinatorial testing as a coverage criterion, whose goal is to ensure that the set of tested scenarios can cover all combinations for arbitrary~$k$ categories. 
To increase $k$-way coverage while avoiding certain scenario combinations, the abstract scenario generation problem is reduced to a mixed-integer linear programming (MILP) problem, with the objective being maximally increasing the coverage. The encoding is borrowed from our earlier work for testing ML-based systems~\cite{cheng2018quantitative}. A simple illustration of the 2-way combinatorial testing can be found in Figure~\ref{Figure2Projection.coverage}.  First, ComOpT builds tables for three categories mentioned earlier, i.e.,  $C_1 = \sig{weather}$,   $C_2 = \sig{road}$, and $C_3 = \sig{ego-action}$. A table is associated with every pair of categories; therefore a total of three ${{3}\choose{2}}$ categories. With a particular combination being impossible (in this simple example, it is stated that it is impossible to do a left turn when the road is straight), the total number of empty cells to be filled equals~$20$ ($C_1,C_2: 9$; $C_2,C_3: 5$; $C_1,C_3:6$). Given that there exists the first abstract scenario  $\langle \sig{sunny}, \sig{drive-straight}, \sig{straight}\rangle$, the MILP will propose the second scenario   $\langle \sig{rainy}, \sig{u-turn}, \sig{T-shaped}\rangle$ that can maximally increase the metric by filling~$3$ cells. When all cells are filled, we have achieved a \emph{relative form of completeness} where every category pair is covered. Observe that two abstract scenarios in Figure~\ref{Figure2Projection.coverage} have very different characteristics. By proceeding with such methods, the testing is truly driven by coverage, and we can guarantee that \emph{every abstract scenario is different from previously generated ones at least in one category}, thereby guaranteeing  diversity.

%\vspace{1mm}
Finally, readers may raise concerns about generating complex routes via an axiomatic approach. Nevertheless, for any complex route within the ODD, one can always view it as a \emph{concatenation of multiple simple sub-journeys} with each sub-journey  to cross a junction or drive along a lane segment (without intersection). Therefore, when ComOpT performs systematic testing, it just considers driving scenarios with various initial configurations under a single junction (e.g., \sig{T-shaped}, \sig{roundabout}) or a lane segment without intersection. 

\subsection{Concrete scenario generation}

From a synthesized  abstract scenario,  ComOpT generates a concrete scenario by querying the map and the simulator. Detailed actions include finding a sub-map that satisfies all geographical conditions, locating the \emph{ego vehicle} (here referring to the vehicle being controlled by the autonomous driving software stack), configuring \emph{Non-Player Character (NPC)} vehicles  and pedestrians, and setting up simulation parameters such as weather. 

\vspace{1mm}

\subsubsection{Sub-map finding based on semantic information}

\begin{figure}[t]
\includegraphics[width=0.49\textwidth]{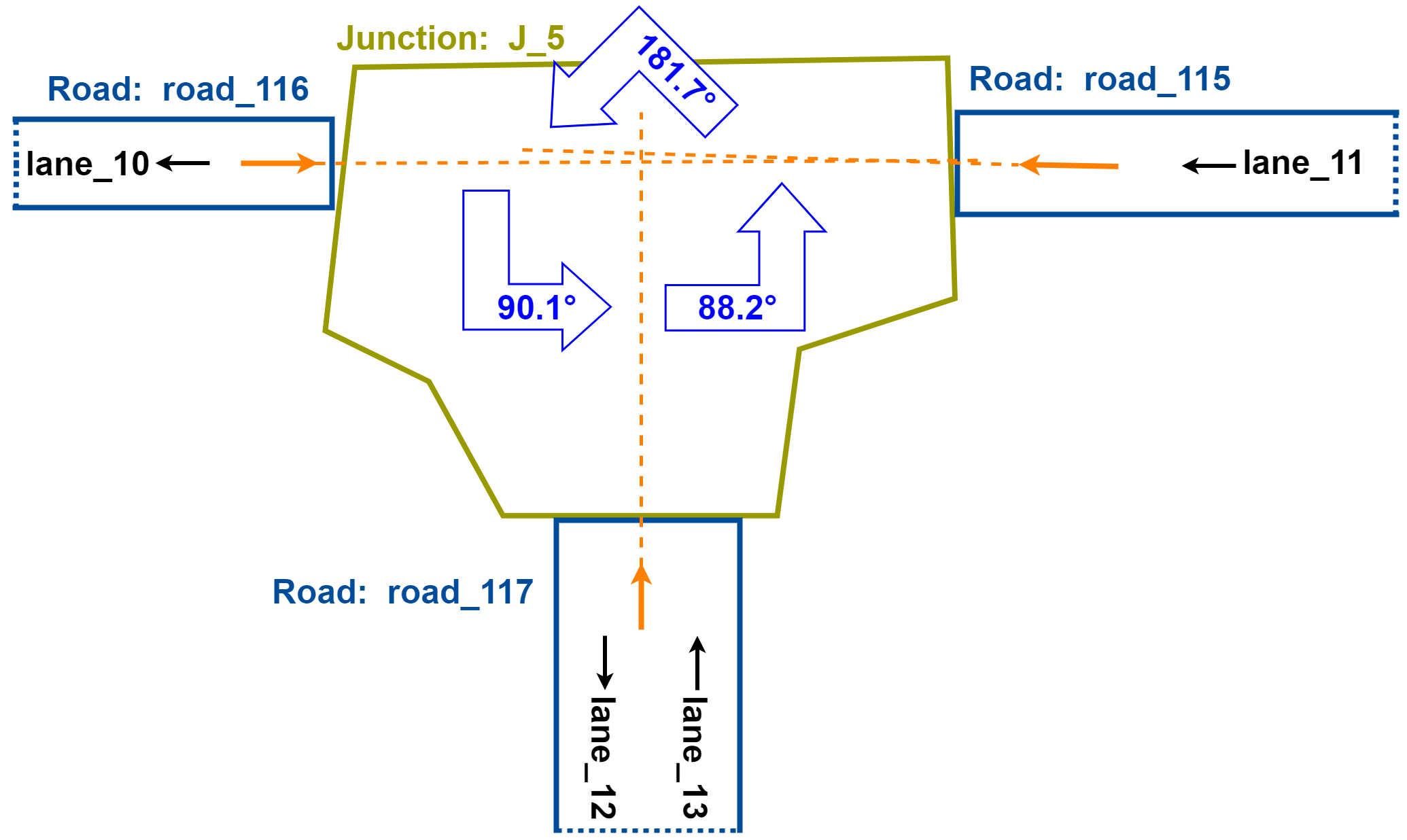}
\centering
\vspace{-5mm}
\caption{Example of road angles in a junction}
\label{fig:tshaped}
\end{figure}

For a specific road structure matching the semantic information, such as \sig{T-shaped} junction, ComOpT needs to search for the corresponding sub-map in the map and assign all agents accordingly. Due to space limits,  we only highlight how a \sig{T-shaped} junction is discovered using the junction example in Figure~\ref{fig:tshaped}. 
We refer readers to the ComOpT documentation (in the source tree: \texttt{scripts/comopt/map\_parse/README.md}) for details on identifying other road structures.

First, for a given junction, ComOpT considers the number of related roads. Road structures with the same number of related roads are allocated into the same group. In Figure~\ref{fig:tshaped}, three roads connect to junction~$\sig{J\_5}$. Then ComOpT computes the angles between related roads. The angle sequence of adjacent roads is used to match the unique angle range sequences of each road structure class. In Figure~\ref{fig:tshaped}, roads with identifiers $\sig{road\_115}$, $\sig{road\_116}$, and $\sig{road\_117}$ form angles of $181.7$, $90.1$, and $88.2$ degrees. The computed degrees match the specification of a \sig{T-shaped} junction, where for an ideal \sig{T-shaped} junction, the incoming roads should form angles of $180$, $90$, and $90$ degrees. Junction $\sig{J\_5}$ will not be categorized as a \sig{Y-shaped} junction; as for a \sig{Y-shaped} junction, incoming roads should form angles around~$120$, $120$, and~$120$ degrees.

\begin{figure}[t]
\includegraphics[width=0.49\textwidth]{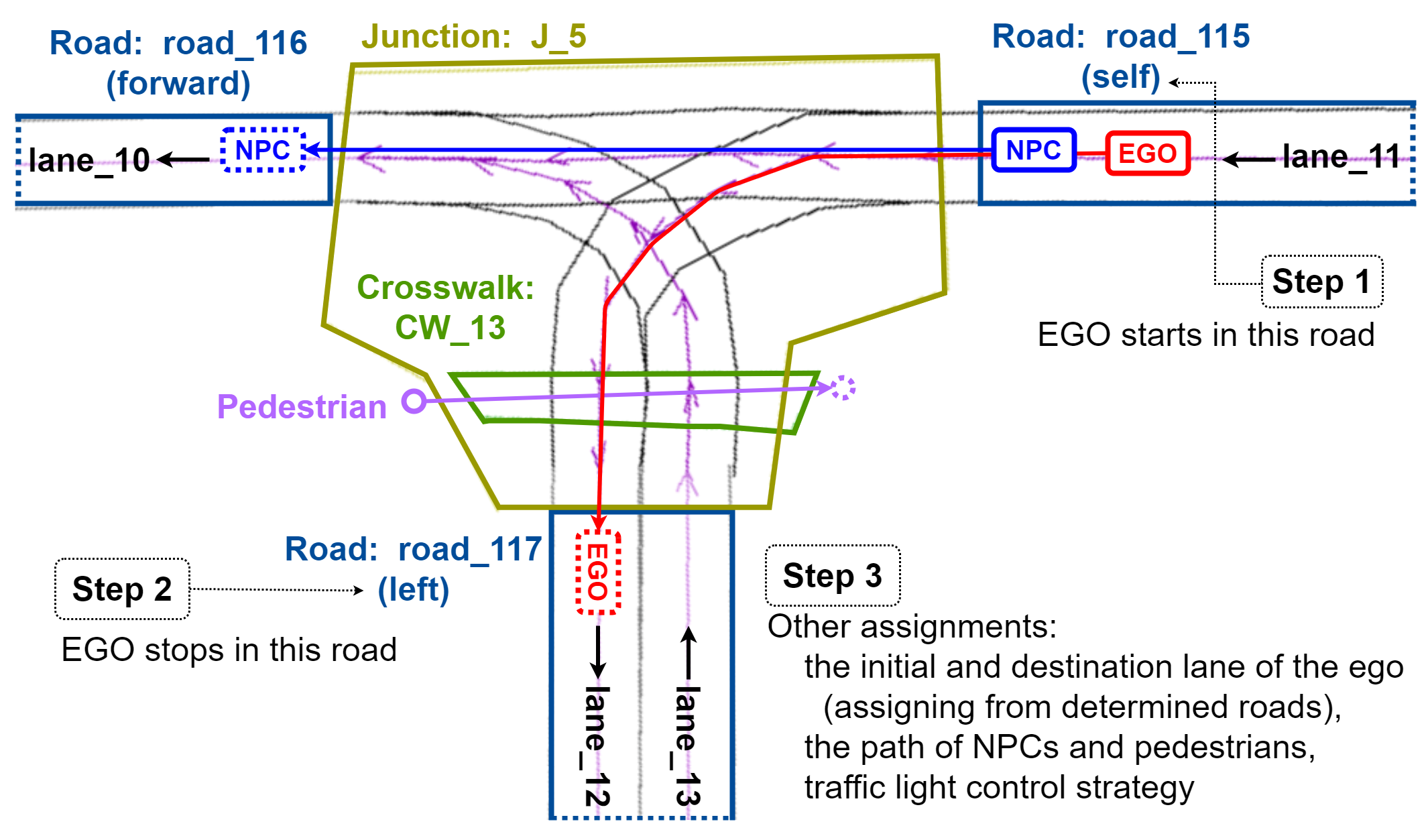}
\centering
\vspace{-5mm}
\caption{Example of generating concrete scenario from map data}
\label{fig:mapexample}
\end{figure}

\vspace{1mm}
\subsubsection{Concrete scenario generation outline}

To demonstrate how ComOpT generates concrete scenarios with map data, we consider the example in Figure~\ref{fig:mapexample} which is a detailed visualization for the junction \sig{J\_5}.  The detailed process is as follows:
\begin{itemize}
% Step1:
    \item \emph{Step 1 - assigning the initial road for the ego vehicle}: We first decide the initial road segment for the ego vehicle, such that the specified ego behavior in an abstract scenario is legal and feasible (without collision with other agents). For example, the initial road where the ego vehicle starts its journey in the example is marked as ``self" (\sig{road\_115}).

% Step2:
    \item \emph{Step 2 - assigning the destination road for the ego vehicle}: The choice for the destination road of the ego vehicle considers the road structure and the behavior of the ego vehicle. For example, consider if the ego vehicle should perform a left turn, as indicated by the abstract scenario. The road marked with \sig{road\_117} is marked as left, as it is in the left of the located road of the ego vehicle. Therefore, we choose a feasible position on \sig{road\_117}  as its destination.

% Step3:
    \item \emph{Step 3 - other assignments}: Finally, ComOpT considers other road-related configurations associated with the scenario, such as the path of NPCs and pedestrians and the traffic light control strategy.
\end{itemize}

\begin{figure*}[t]
\includegraphics[width=0.9\textwidth]{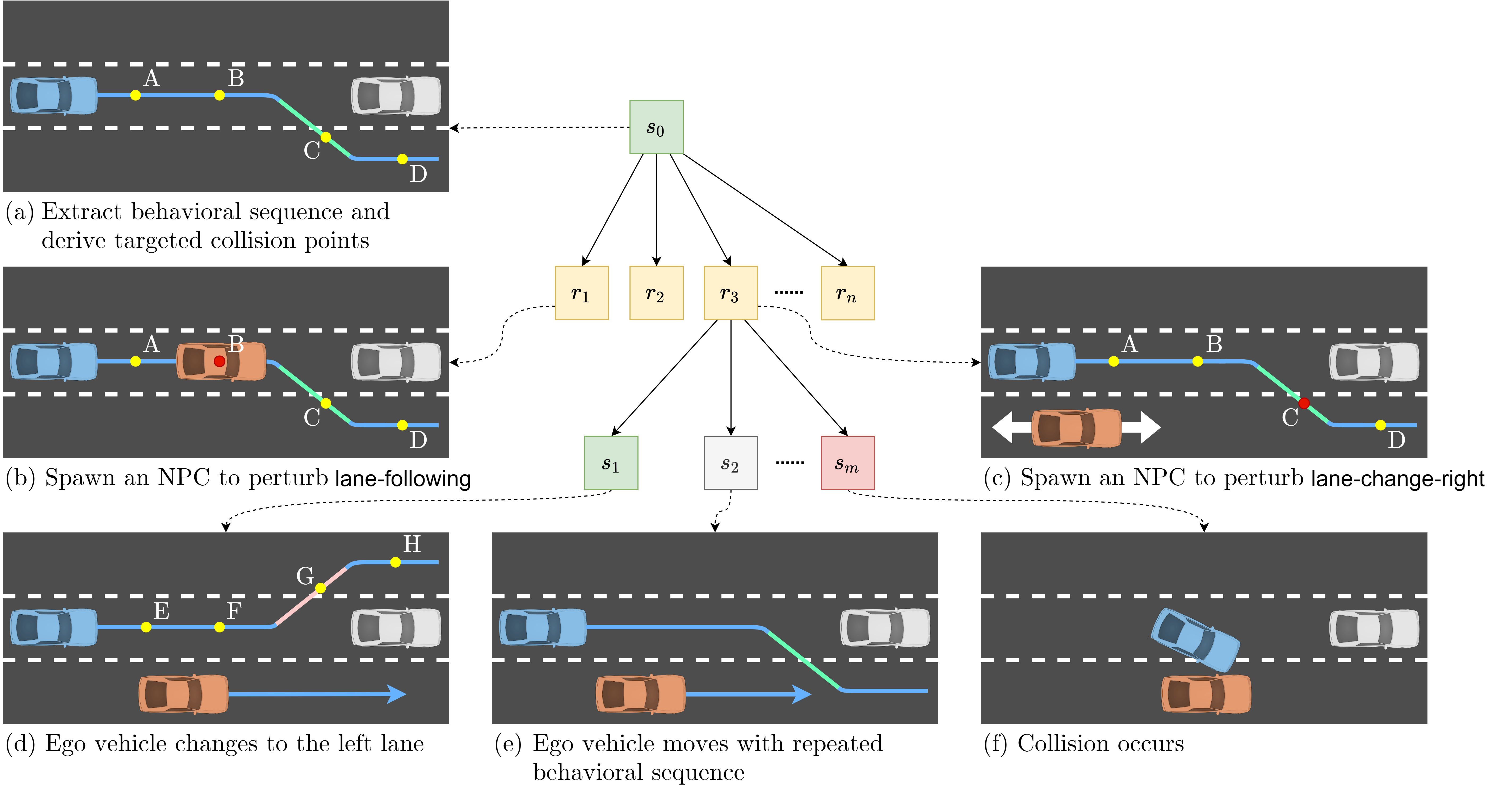}
\centering
\vspace{-2mm}
\caption{Illustrating the process of agent spawning}
\label{fig:agentspawning}
\vspace{-5mm}
\end{figure*}

\subsubsection{Non-road related parameter instantiation}

For other parameters, the instantiation from an abstract scenario to a concrete scenario requires a mapping process to the simulated world. As an example, in the parameter mapping file, it is specified that the weather ``\sig{cloudy}" stated in the abstract scenario is mapped to the LGSVL simulator with the following parameter range: 

\begin{itemize}
    \item ``cloudiness": $[0.3, 1.0]$,
    \item  ``rain": $[0.0, 0.1]$,
    \item ``wetness" : $[0.0, 0.3]$,
    \item ``fog" : $[0.0, 0.3]$
\end{itemize}

Therefore, when instantiating concrete scenarios, one can either select the medium value or assign a random value within the range. ComOpT predominantly uses random assignment, but for parameters to be perturbed in later scenario variation (e.g., \sig{number-of-vehicles}), ComOpT carefully selects the value to avoid being on the boundary.

\subsection{KPI for evaluating the quality of the scenario}

Once when a concrete scenario is created, ComOpT runs the scenario and uses a set of metrics to evaluate if  an undesired situation occurs. The set of undesired situations considered by ComOpT includes (a) \emph{collision or being very close}, (b) \emph{uncomfortable brake or sudden acceleration}, (c) \emph{lateral jerk}, (d) \emph{route deviation}, and finally (e) \emph{ignoring traffic signals}. Note that for route deviation, it includes events such as driving out-of-road (e.g., onto the sidewalk) as well as inabilities to make simple right turns when the destination is simply placed at the right of the intersection.

\subsection{Local scenario variation via agent spawning}\label{sub.sub.local.scenario.variation}

When a concrete scenario does not demonstrate undesired behavior, ComOpT introduces a new scenario perturbation technique called \emph{agent spawning} to find scenario variations that can make the resulting simulation demonstrating undesired behavior. Initially, the concrete scenario may have some agents (e.g., pedestrians or other road vehicles). The maximum number of additional agents that can be added is constrained by the equivalence class specified in the abstract-to-concrete mapping information. E.g., for vehicle density being \sig{mild}, in the concrete scenario, it is possible to have the number of vehicles within range $[3, 6]$. Therefore, if the current concrete scenario has only~$4$ vehicles, it is still possible to spawn~$2$ additional vehicles while the scenario is still within the equivalence class defined by the abstract scenario. 

The premise of applying agent spawning requires analyzing the \emph{timed trace} over a concrete scenario. A timed trace represents the state of every agent (ego vehicle, NPC, traffic light, etc) under a predefined time granularity (e.g., $0.1$ second) recorded in the simulation. 
The purpose of analyzing the timed trace contains two goals: (a) extract the high-level behavioral information of the ego vehicle, and based on the high-level behavioral information, (b) decide how to inject a new NPC in the existing scenario that may lead to a collision. In the following, we detail the underlying workflow.   

\vspace{1mm}
\subsubsection{Extracting the behavioral sequence}

ComOpT first analyzes the timed trace and extracts the \emph{behavioral sequence} of the ego vehicle, which is a summary of all intermediate actions conducted by the ego vehicle when moving from the source waypoint to the destination waypoint. 
Consider the example in Figure~\ref{fig:agentspawning}a. It shows the trace of the ego vehicle in a simulated episode from a concrete scenario. Intuitively, the ego vehicle first follows the lane, then performs a lane change, and finally continues to follow the new lane.  The behavioral sequence can thus be summarized as $\langle \sig{lane-following}, \sig{lane-change-right}, \sig{lane-following}\rangle$. In ComOpT, each pattern in the alphabet (for building the behavioral sequence) has a precise interpretation to be mapped to a segment of the timed trace. For example, 
\begin{itemize}
    \item the pattern ``\sig{lane-change-right}" matches a segment of the timed trace starting when the bounding box of the ego vehicle intersects with the right lane separator until the bounding box is fully contained in the right adjacent lane being in the same driving direction, and 
    
    \item the pattern ``\sig{encroaching-change-left}" is similar to \sig{lane-change-right} with the difference that the vehicle moves to the adjacent left lane being in the opposite driving direction. 
\end{itemize}

\vspace{1mm}
\subsubsection{Extracting the targeted collision points}

Subsequently, ComOpT extracts the \emph{targeted collision points} out of the behavioral sequence, where we expect the introduced NPC to collide with the ego vehicle at one of the points. 
Consider again Figure~\ref{fig:agentspawning}a, where two points~$A$ and~$B$ are selected for the first segment \sig{lane-following}, and point~$C$ is selected for the segment \sig{lane-change-right}. Again, ComOpT has a predefined rule in terms of extracting targeted collision points. For instance, for the segment \sig{lane-change-right}, ComOpT always extracts from the timed trace the position where the center of the ego vehicle first crosses the lane boundary.

\vspace{1mm}
\subsubsection{Altering the scenario by adding one new  NPC}

Subsequently ComOpT decides, for each targeted collision point and its associated behavioral pattern, possible ways for introducing an NPC to induce a collision. We use again the scenario in Figure~\ref{fig:agentspawning}a to explain the concept. 

\begin{itemize}
    \item For targeted collision point $C$ where the ego vehicle performs a \sig{lane-change-right}, a natural way of introducing an NPC is to \emph{allocate it on the adjacent right lane}. Subsequently, control the configuration such that when the NPC drives along the lane, at time~$t$ when the ego vehicle reaches point~$C$, the position of the NPC is also very close to~$C$. This leads to the new scenario as demonstrated in  Figure~\ref{fig:agentspawning}c.  
    
    \item For targeted collision point $B$ where the ego vehicle performs a \sig{lane-following}, a natural way of introducing an NPC is to \emph{allocate it on the same lane} with a configuration such as abrupt braking. This leads to the new scenario as demonstrated in  Figure~\ref{fig:agentspawning}b.  
\end{itemize}

While for each pattern (e.g., $\sig{lane-following}$) there is a fixed set of NPC control strategies (e.g., braking or being stationary), how to derive the concrete configuration remains to be solved. For simple control strategies such as setting the NPC to be stationary, deriving the configuration is trivial. For other cases, one needs to utilize physics to infer the possible configuration. 
For example, consider the scenario in Figure~\ref{fig:agentspawning}c where ComOpT plans to spawn an NPC to drive along the adjacent lane. Provided that the NPC shall be close to position~$C$ at time~$t$, if the NPC drives under a fixed velocity~$v$, then it should be placed initially at a position that is approximately~$vt$ meters away from point~$C$. In the implementation, the distance to point~$C$ is relaxed by a set $[vt-\Delta, vt+\Delta]$, with~$\Delta$ being a constant to allow considering nearby starting positions.

\vspace{1mm}
\subsubsection {Scenario testing via optimization-based parameter search} 

Summarizing the process until now, given a concrete scenario, for each targeted collision point, ComOpT derives a \emph{parameterized scenario} where the parameter is the initial distance between the newly spawned NPC and the targeted collision point. 
Finally, ComOpT performs a systematic search over the parameterized scenario. One can adopt multiple strategies such as randomization or uniform sampling to derive concrete parameters, in order to find scenarios with the ego vehicle demonstrating undesired behavior. However, testing a scenario is computationally expensive: it requires executing the autonomous driving stack in the simulation environment. Therefore, we have used an \emph{optimization-based approach} to guide the finding of suitable parameters that lead to a collision.   

The optimization target is to reach the situation where (1) the newly introduced NPC first reaches the targeted collision point, thereby generating possibilities for the ego vehicle to hit the NPC. Meanwhile, (2) the time when the introduced NPC reaches the targeted collision point should ideally be slightly earlier than the ego vehicle.

Therefore, if the time when the NPC drives through the targeted collision point is substantially earlier than the time when the ego vehicle reaches there, ComOpT increases the  distance between the NPC's initial location and the targeted collision point. On the contrary, when the NPC arrives at the targeted collision point much later than the ego vehicle, ComOpT decreases the initial distance. Due to space limits, we refer readers to the source code for concrete configurations such as step size. 

\vspace{1mm}

\subsubsection{Managing the ordering of scenarios to be tested, and termination} 

In previous subsections, we detail how to derive a new scenario to be tested, under a unique targeted collision point. However, as shown in Figure~\ref{fig:agentspawning}a, there are multiple targeted collision points ($A, B, C, D$) to be considered. To this end, one requires a \emph{meta-level search strategy} to manage the ordering on the scenarios to be tested, as well as setting up proper termination criteria. We exemplify the underlying meta-level search strategy using Figure~\ref{fig:agentspawning}. 

First, ComOpT maintains a \emph{priority queue} over the targeted collision points, where the priority is based on the associated behavioral pattern of the point. For example, \sig{lane-change-right} has higher priority over \sig{lane-following}. Therefore, in Figure~\ref{fig:agentspawning}a, the targeted collision point~$C$ is explored first, leading to the parameterized scenario in Figure~\ref{fig:agentspawning}c. By varying the initial location of the NPC and running the simulation, one of the situations may occur:
\begin{itemize}
    \item The ego vehicle collides with other NPCs, as demonstrated in Figure~\ref{fig:agentspawning}f. Then the search terminates. 
    \item The ego vehicle generates the same behavioral sequence $\langle \sig{lane-following}, \sig{lane-change-right}, \sig{lane-following}\rangle$, as demonstrated in Figure~\ref{fig:agentspawning}e. As ComOpT stores all visited behavioral sequences that have been used, this scenario is not further explored. 
    \item The ego vehicle generates a new behavioral sequence $\langle \sig{lane-following}, \sig{lane-change-left}, \sig{lane-following}\rangle$, as demonstrated in Figure~\ref{fig:agentspawning}d. This scenario will be further explored. Therefore, targeted collision points $E, F, G$, and $H$ will be added to the existing priority queue containing~$A$, $B$, and~$D$. ComOpT then selects point~$G$ to be further explored, due to its associated pattern \sig{lane-change-left} having higher priority. 
\end{itemize}

Finally, to ensure termination, ComOpT sets a fixed budget on the number of simulations to be executed whenever local scenario perturbation algorithm is triggered.

%%%%%%%%%%%%%%%%
\section{Evaluation on Baidu Apollo and LGSVL}

This section details the result evaluated on the Baidu Apollo 6.0 AD stack (\texttt{master} branch taken on 2021.04.14) and the LGSVL simulator (version \text{2021.01}). The configuration reported in this paper is based on the version that participated in the 2021 IEEE Autonomous Vehicle Testing Challenge.  We refer readers to the following YouTube channel for a summary of results, including (1) a two-minute teaser video highlighting the techniques and some undesired scenarios, and (2) additional~$40$ problematic AD driving scenarios discovered by ComOpT. 

\vspace{2mm}
{\small\url{https://www.youtube.com/playlist?list=PL6ii4xJXGd8L3bn9pWuchuKBPb0R5JgJ7}}

\vspace{2mm}

For the version used in the competition, ComOpT is set with a default configuration to generate the first~$15$ abstract scenarios (equivalence classes) that maximally increase coverage governed by $2$-way combinatorial testing. We assume that the ODD is defined using the San Francisco map as supported by the LGSVL simulator. Therefore, the atomic map geometry only contains straight lanes, T-way junctions, and 4-way intersections. We need to explicitly specify constraints such as ``no roundabouts" to prevent the abstract scenario generator from creating such candidates. 

%\vspace{2mm}
Table~\ref{tab:total} highlights the statistical summary of the generated scenarios. Every column lists the number of such scenarios. The scenario set \texttt{base} has $45$ concrete scenarios, as each abstract scenario ($15$ in total) is instantiated~$3$ times to build concrete scenarios. The case of \texttt{perturbed} considers the generated scenarios by perturbing (detailed in Sec.~\ref{sub.sub.local.scenario.variation}) the concrete scenarios in the \texttt{base}. The problems on \textit{too-close} and \textit{collision} are regarded as safety-critical. The others are undesirable but regarded as performance issues. 
A scenario may contain both safety-critical and performance issues. 
From Table~\ref{tab:total}, we observe that our local scenario perturbation technique, at least within the experiment, is effective in uncovering safety-critical scenarios. We observe that there are only two types of collision accidents between vehicles collected from the simulation results: \emph{rear collision} and \emph{lateral collision}. The reason is that  the San Francisco  map of offered by LGSVL only covers a part of the city. Roads within this part of the city does not have dashed yellow lines, and overtaking by borrowing the lane in the opposite direction is not allowed in Apollo. In such a case, \emph{head-to-head collisions cannot happen} unless agents are educated to intentionally violate the traffic law. 

\begin{table}[tp]
\caption{Statistics on the generated scenarios}
    \label{tab:total}
    \centering
    \begin{tabular}{c|c|c|c|c}\hline
         types &total & problematic &  safety-critical  &   performance   \\\hline
         \texttt{base} & 45 & 39 &5 &39\\
         \texttt{perturbed} & 273 & 243 & 130&225\\\hline
    \end{tabular}
    
\end{table}

Studying the root cause of undesired behaviors, we realize that Apollo may (1) fail to make trivial right turns, as it does not perform a lane change first, (2) fail to keep a safe distance, (3) have frequent interleaving of acceleration and deceleration, even when no external traffic signal exists, (4) violate the traffic rules such as running over red lights when there is still sufficient space to stop, and finally (5) run into trouble when inconsistencies exist  between the map in the LGSVL simulator and the map internally stored in Apollo.

\section{Summary}

ComOpT is our initial step towards a vision where testing autonomous driving systems can be systematically approached with scientific rigor. Within ComOpT, we designed and implemented a coverage-driven, two-layered approach that guarantees to maintain abstract scenario diversity while capable of performing local scenario perturbation. 

ComOpT still has many potentials to be improved and matured, such as integrating the methods in the literature for generating scenarios, or designing other traffic agents beyond using the NPC agents made available by the simulators, or providing proper interfaces to other autonomous driving SW stacks such as Autoware.Auto~\cite{AUTOWARE}. 

\bibliographystyle{IEEEtran}

\end{document}